\documentclass{article}

\usepackage{microtype}
\usepackage{graphicx}
\usepackage{tabularx}
\usepackage{subcaption}
\usepackage{booktabs} 
\usepackage{xcolor}   
\usepackage{amssymb}  
\usepackage{placeins}

\usepackage{hyperref}

\usepackage[preprint]{icml2026}

\usepackage{amsmath}
\usepackage{mathtools}
\usepackage{amsthm}
\usepackage{enumitem}
\usepackage{bm}
\usepackage[capitalize,noabbrev]{cleveref}

\colorlet{taskletsgray}{gray!12}
\definecolor{taskletsblue}{HTML}{EBF4FC}
\definecolor{taskletsgreen}{HTML}{EFF8EA}
\definecolor{taskletsteal}{HTML}{E7F9F6}
\definecolor{taskletscoral}{HTML}{FCEFEB}
\definecolor{taskletslavender}{HTML}{F0F0FE}
\definecolor{taskletsyellow}{HTML}{FFFBE4}
\definecolor{taskletspeach}{HTML}{FFF3E5}

\newcommand{\tokenlyfont}[1]{\colorbox{taskletsgray}{\texttt{#1}}}

\usepackage[listings]{tcolorbox}

\newtcblisting{taskexample}[1][taskletsyellow]{
  listing only,
  listing options={
    basicstyle=\small\ttfamily,
    columns=fullflexible,
    keepspaces=true
  },
  colback=#1,
  colframe=#1,
  boxrule=0pt,
  arc=0pt,
  hbox,
  tcbox width=auto, 
  boxsep=0pt,
  left=6pt,
  right=6pt,
  top=5pt,
  bottom=5pt,
  before skip=1.0em,
  after skip=1.0em
}

\newcommand{\taskletstile}[2]{%
  \colorbox{#1}{%
    \kern-\fboxsep\vphantom{Tasklets}#2\kern-\fboxsep}%
}

\DeclareRobustCommand{\taskletslogo}{%
  \mbox{\ttfamily%
    \taskletstile{taskletscoral}{\kern\fboxsep T}%
    \taskletstile{taskletsgreen}{a}%
    \taskletstile{taskletsblue}{s}%
    \taskletstile{taskletspeach}{k}%
    \taskletstile{taskletsteal}{l}%
    \taskletstile{taskletslavender}{e}%
    \taskletstile{taskletsyellow}{ts\kern\fboxsep}%
  }%
}

\theoremstyle{plain}

\theoremstyle{definition}

\theoremstyle{remark}

\usepackage[textsize=tiny]{todonotes}

\icmltitlerunning{Tasklets: A Unit-Test Benchmark for Language Model Architectures}

\begin{document}

\twocolumn[
  \icmltitle{\tokenlyfont{Tasklets}: A Unit-Test Benchmark for Language Model Architectures}



  \icmlsetsymbol{equal}{*}

    \begin{icmlauthorlist}
      \icmlauthor{Zayd M. K. Zuhri}{equal,mila}
      \icmlauthor{Munachiso Samuel Nwadike}{equal,mbzuai,riken}
      \icmlauthor{Kentaro Inui}{mbzuai,riken,tohoku}
      \icmlauthor{Alham Fikri Aji}{mbzuai}
    \end{icmlauthorlist}
    
    \icmlaffiliation{mila}{Mila, Canada}
    \icmlaffiliation{mbzuai}{MBZUAI, UAE}
    \icmlaffiliation{riken}{RIKEN AIP, Japan}
    \icmlaffiliation{tohoku}{Tohoku University, Japan}

    \icmlcorrespondingauthor{Zayd M. K. Zuhri}{zayd.zuhri@mila.quebec}
    \icmlcorrespondingauthor{Munachiso Samuel Nwadike}{munachiso.nwadike@mbzuai.ac.ae}
    
  \icmlkeywords{Machine Learning, ICML}

  \vskip 0.3in
]



\printAffiliationsAndNotice{\icmlEqualContribution}

\begin{abstract}
  We introduce \tokenlyfont{Tasklets}, a benchmark of small synthetic tasks for evaluating the computational capabilities of language model sequence-mixing layers. These tasks probe foundational capabilities such as recall, sequence manipulation, and state tracking. Such tasks are widely used in architecture research, but differences in task selection, formulation, and evaluation make comparisons across studies difficult. \tokenlyfont{Tasklets} brings these tasks together, organizing them by the capabilities they test and providing a shared setup for evaluating how different architectural layer types learn fundamental computations. We compare a broad range of layer types through controlled experiments with small models. Our task implementations and evaluation code support reproducible, low-cost testing of new architectural proposals: \url{https://github.com/zaydzuhri/flame/tree/tasklets} .
\end{abstract}

\section{Introduction}
\label{sec:introduction}

Modern language models are built from stacks of blocks, within which are individual computational layers \cite{nvidia2026nemotron35bolt,qiu2026qwen38next,kimiteam2026kimik3,deepseekai2026v41flash,gemmateam2026gemma4}. These include channel-mixing layers (such as multi-layer perceptrons or mixture-of-experts layers) and sequence-mixing layers (also referred to as token-mixing layers, such as multi-head attention or Mamba layers) \cite{kimiteam2025kimilinearexpressiveefficient,yang2025gated,lahoti2026mamba3,poli2024mechanistic}. In this work, we propose a unit-test for sequence-mixing layers.

\begin{figure}[t]
\begin{center}
\includegraphics[width=\linewidth]{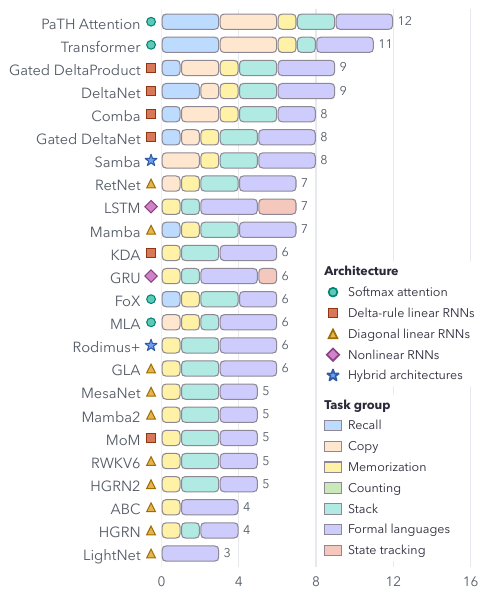}
\end{center}
\caption{Number of tasks solved by each architecture on \tokenlyfont{Tasklets}, where a task counts as solved at accuracy $\geq 0.9$. Bar segments show the capability group of each solved task, and the symbol beside each name marks the architecture family. Architectures are sorted by tasks solved, with ties broken by mean accuracy. Per-task accuracies are shown in Figure~\ref{fig:main_heatmap}.}
\label{fig:tasks_solved}
\end{figure}

Different layer types support different computational capabilities, which can be characterised as architectural expressivity \cite{shakerinava2026expressive,merrill2026parallelizable,illusionofstate}. However, in practice, determining which layers to use and how to combine them is often an experimental and resource intensive process. Modest design changes may potentially require costly training runs and long evaluation cycles. Researchers often address this issue by leveraging \underline{synthetic tasks} \cite{lstm6795963,poli2024mechanistic,arora2023zoologymeasuringimprovingrecall,jelassi2024repeat,kimiteam2025kimilinearexpressiveefficient,yang2025path}.

These synthetic tasks isolate specific computational capabilities, such as selective copying, state tracking, or recall, and have a long history in sequence modelling, including early work on LSTMs \cite{lstm6795963}, external-memory architectures \cite{graves2014neuralturingmachines,learningtotransduce}, and fast weights \cite{ba2016usingfastweightsattend}.

Synthetic evaluation is often fragmented across studies. For instance, the same task may appear under different names, such as the task referred to as sequence reversal \cite{learningtotransduce}, reverse \cite{cai2025extrapolation}, and palindrome \cite{kimiteam2025kimilinearexpressiveefficient}. Furthermore, studies may use different tasks to test the same underlying capability. For example, \citet{jelassi2024repeat} test copying by asking models to reproduce an entire sequence, while \citet{gu2024mamba} test the ability to copy while ignoring distractors, and \citet{cai2025extrapolation} focus on the ability to return the input in reverse order. Recall evaluations similarly differ in whether models must retrieve a value for a single query \cite{arora2025mechanistic}, answer multiple queries \cite{arora2023zoologymeasuringimprovingrecall}, or handle associations that span multiple tokens or are embedded among distractors \cite{poli2024mechanistic}.

When different architectures are evaluated on marginally different tasks intended to test the same capability, variations in task demands complicate direct comparisons. Success on one task also does not establish success across the broader capability. Bringing these tasks and their variants together is therefore valuable. While prior work \cite{poli2024mechanistic} proposes such a unified benchmark in the context of hybrid architectures, one is still needed for sequence-mixing layers.

We thus introduce \tokenlyfont{Tasklets}, an open-source, computationally inexpensive synthetic unit-test benchmark for sequence-mixing layers. We evaluate a broad range of layer types, examining task performance and generalisation, and provide task implementations and evaluation code to support reproducible testing of new architectural proposals.

\section{Task Sourcing}
\label{sec:our_tasks}

\tokenlyfont{Tasklets} exclusively aggregates tasks previously utilised in papers proposing novel sequence architectures. It groups them by the 7  capabilities they test as follows:

\textbf{Recall.} \citet{graves2014neuralturingmachines} used associative recall to evaluate Neural Turing Machines. Our multi-query recall variant follows \citet{arora2023zoologymeasuringimprovingrecall}, who introduced the task to compare attention with gated convolutions. Our fuzzy and noisy recall variants follow \citet{poli2024mechanistic}. \citet{jelassi2024repeat} refer to fuzzy recall as $n$-gram lookup, which they used to compare Transformers, Mamba, and LSTMs.

\textbf{Copying.} Like recall, an early formulation of copying appears in \citet{graves2014neuralturingmachines}. Our full copy and selective copy variants follow \citet{poli2024mechanistic}, who used selective copying to guide hybrid architecture design. Selective copy is referred to as denoising in \citet{jing2018gated} and is used to evaluate their proposed gated orthogonal recurrent units (GORUs). Our reverse copy variant follows the sequence reversal task used by \citet{learningtotransduce} to evaluate memory-augmented LSTMs. This task is also referred to as reverse by \citet{cai2025extrapolation}, who studied Transformer length generalisation, and palindrome by \citet{kimiteam2025kimilinearexpressiveefficient}, who evaluated Kimi Delta Attention (KDA). 

\textbf{Memorisation.} We adopt the fixed key-value memorisation task used by \citet{poli2024mechanistic}.

\textbf{Counting.} We expand the majority counting used to evaluate xLSTMs \cite{beck2024xlstm}, to return counts for all queried tokens. \citet{chang2025language} used this to study length generalisation across RNNs, Transformers, SSMs, and RWKV. We also include frequency based sorting following \citet{martins-etal-2022-former}, who used it to assess long term information retention in their proposed \(\infty\)-former.

\textbf{Stack operations.} \citet{deletang2023neural} used stack manipulation to investigate length generalisation across neural architectures, and \citet{beck2024xlstm} subsequently used it to evaluate xLSTM. Our single-stack variant adapts this task by starting with an empty stack and predicting the value returned at each pop. We also include the multi-stack variant following \citet{kimiteam2025kimilinearexpressiveefficient}.

\textbf{Formal languages.} Flip-flop language modelling was introduced by \citet{liu2023exposing} to investigate attention glitches in Transformers and subsequently used by \citet{yang2025path} to evaluate PaTH Attention. \citet{joulin2015inferring} used \(a^n b^n c^n\) sequence prediction to evaluate their stack-augmented recurrent networks, while \citet{suzgun2019memory} used Dyck languages to evaluate their proposed memory-augmented recurrent architectures. \(a^n b^n c^n\) and Dyck languages are also discussed in \citet{deletang2023neural}.
 
\textbf{State tracking.} \citet{illusionofstate} used modular addition and permutation composition to compare state tracking in S4, Mamba, and their proposed IDS4. We adapt their modular addition task from \(\mathbb{Z}_{60}\) to \(\mathbb{Z}_{10}\) and retain their \(A_5\) permutation composition task. Modular arithmetic was also used by \citet{beck2024xlstm} to evaluate xLSTM.

We describe each task in Section~\ref{sec:benchmark_details}.

\section{\tokenlyfont{Tasklets} Benchmark}
\label{sec:benchmark_details}

We propose a unit-test benchmark for sequence-mixing layers comprising 16 synthetic tasks across seven capability categories. Below, we specify each task's format and data-generation settings.

For each task, we generate 100,000 training examples and 3,000 test examples, with the vocabulary size set to 292.

In the examples below, `\texttt{x}' denotes the input sequence and `\texttt{y}' denotes the target outputs. Underscores `\texttt{\_}' in \texttt{y} indicate masked positions excluded from the training loss and accuracy calculation i.e they are neither part of the \texttt{y} tokens that count towards cross entropy loss nor towards evaluation performance. Each predicted \texttt{y} token is predicted at the \texttt{x} position immediately preceding it.

For associative recall and counting, the separator token `\texttt{|}' separates the context from the query portion. For copying and sorting, it separates the source sequence from the output sequence. For language recognition, it marks the end of the sequence to be classified. In memorisation, it separates successive entries of key-values.

\subsection{Recall} 
\label{sec:recall_task_explanation}

\subsubsection{Multi-Query Recall}

A model may need to remember several facts introduced in a passage and retrieve different facts at different points. In multi-query recall, the input to the layer is a sequence of key-value pairs, where each key and each value is a single token. This sequence is followed by several queries asking for the values associated with previously presented keys. For example:

\begin{taskexample}[taskletsblue]
x: 1 6 4 7 3 8 5 9 2 6 | 4 7 3 8 2 6
y: _ _ _ _ _ _ _ _ _ _ _ _ 7 _ 8 _ 6
\end{taskexample}
Above, the context specifies that key \texttt{4} has value \texttt{7}, key \texttt{3} has value \texttt{8}, and key \texttt{2} has value \texttt{6}. When queried with \texttt{4}, the model must return \texttt{7}. When queried with \texttt{3}, it must return \texttt{8}. When queried with \texttt{2}, it must return \texttt{6}. Answering these queries requires retaining and retrieving several associations from the same context.

In our experiments, each sequence contains a context of 128 tokens followed by 32 queries. Including the separator, queries, and associated values, each complete sequence contains 193 tokens.

\subsubsection{Noisy Recall}

The context provided to a model may contain unrelated material alongside the information needed to answer later queries. Noisy recall extends our multi-query recall task by inserting distractor tokens from a separate vocabulary into the context. These tokens can appear both between key-value pairs and between a key and its value. For example:

\begin{taskexample}[taskletsblue]
x: 5 4 6 0 7 3 2 8 1 5 9 0 | 4 6 7 2 1 9
y: _ _ _ _ _ _ _ _ _ _ _ _ _ _ 6 _ 2 _ 9
\end{taskexample}

Above, key \texttt{4} has value \texttt{6}, key \texttt{7} has value \texttt{2}, and key \texttt{1} has value \texttt{9}. All other tokens before the \texttt{|} token are distractors. The model must filter out this noise while reading the context and retain the relevant associations for subsequent queries. 

For this task, we use a noise probability of 0.3, a context length of 128 tokens, and 32 queries per sequence. Including the separator, queries, and associated values, each complete sequence contains 193 tokens.



\subsubsection{Fuzzy Recall}  

In language, several neighbouring tokens can together identify a person, an object, or an idea. For instance, a passage might describe the ``wooden front door'' as blue and the ``iron back door'' as green. Remembering these descriptions involves associating each colour with the corresponding phrase. Motivated by this need to associate information with groups of tokens, fuzzy recall extends multi-query recall by allowing each key and each value to consist of one or more consecutive tokens. Their lengths can vary within the same sequence. For example: 

\begin{taskexample}[taskletsblue]
x: 6 4 7 3 8 2 9 5 1 4 9 | 7 3 8 5 1 4 9
y: _ _ _ _ _ _ _ _ _ _ _ _ _ _ 8 _ _ _ 9
\end{taskexample}

Above, key \texttt{6} has value \texttt{4}, and key \texttt{2} has value \texttt{9}. Neither of these keys is queried. The two-token key \texttt{7,3} has value \texttt{8}, and the three-token key \texttt{(5,1,4)} has value 9. When queried with \texttt{7,3}, the model must return \texttt{8}. When queried with \texttt{(5,1,4)}, it must return \texttt{9}. The model must therefore recognise the complete sequence forming each key and retrieve its associated value. Although the values in this example each contain one token, values can also contain several tokens, all of which must be returned in order. For example:

\begin{taskexample}[taskletsblue]
x: 9 2 7 4 8 3 6 5 2 | 9 2 7 4 8 3 6 5 2
y: _ _ _ _ _ _ _ _ _ _ _ _ 7 4 8 _ _ 5 2
\end{taskexample}

In this case, the two-token key \texttt{(9,2)} has the three-token value \texttt{(7,4,8)}, while the two-token key \texttt{(3,6)} has the two-token value \texttt{(5,2)}. When queried with either key, the model must retrieve its complete value and return its tokens in order.
 
We use a window size of 4 and 32 queries per sequence. While the main context contains 128 tokens, the complete sequences have variable lengths once the queries and associated values are included, and are padded on the right.

A key difference between recall and memorisation (which we will study in Section~\ref{sec:memorisation_task_explanation}), is that memorisation uses a fixed mapping learned during training, whereas recall requires retrieving associations provided in the current context. For example, in memorisation, key \texttt{(1,3,5)} always has value \texttt{2}, learned during training. In recall, the same key may have value \texttt{2} in one context and \texttt{8} in another.

\subsection{Copy}
\label{sec:copy_task_explanation}
\subsubsection{Full Copy}
Full copy tests whether a model can reproduce the entire source sequence in its original order after a separator. For example:
\begin{taskexample}[taskletspeach]
x: 4 0 7 2 9 5 1 8 3 | 4 0 7 2 9 5 1 8 3
y: _ _ _ _ _ _ _ _ _ _ 4 0 7 2 9 5 1 8 3
\end{taskexample}
Above, the model must reproduce the sequence before the \texttt{|} token from left to right, preserving every token and its position within the sequence.

The sequences are much longer than in the example above, with each source sequence containing 128 tokens (256 tokens across the source and copied sequences).

\subsubsection{Reverse Copy}
Reverse copy (or palindrome) tests whether a model can reproduce the source sequence in the opposite order, starting with its last token and ending with its first. For example:
\begin{taskexample}[taskletspeach]
x: 8 3 6 1 9 4 2 5 7 | 7 5 2 4 9 1 6 3 8
y: _ _ _ _ _ _ _ _ _ _ 7 5 2 4 9 1 6 3 8
\end{taskexample}

Above, the sequence after the \texttt{|} token is the reverse of the sequence before it.

In our experiments, the source sequence contains 128 tokens (256 tokens across the source and reversed sequences).

\subsubsection{Selective Copy}
In practice, a context may contain distractors alongside the information to be copied. Selective copy tests whether a model can ignore distractor tokens, denoted by \texttt{n}, and reproduce the remaining tokens in their original order. For example:
\begin{taskexample}[taskletspeach]
x: n 7 n n 2 9 n 4 n n 0 6 | 7 2 9 4 0 6
y: _ _ _ _ _ _ _ _ _ _ _ _ _ 7 2 9 4 0 6
\end{taskexample}
Above, the model must ignore every occurrence of \texttt{n} and return \texttt{(7,2,9,4,0,6)}. The number and positions of the distractors do not change the required output.

We use a noise probability of 0.3 to introduce distractors into source sequences of 128 tokens (256 tokens in the complete model input).

\subsection{Memorization}
\label{sec:memorisation_task_explanation}
 
Memorisation tests whether a model can retrieve the exact value associated with a key using a fixed mapping learned during training. The key can be a single token or a collection of tokens. For example:

\begin{taskexample}[taskletsyellow]
x: 8 2 5 9 | 3 7 0 4 | 6 1 8 2 | 9 4 3 7
y: _ _ _ 9 _ _ _ _ 4 _ _ _ _ 2 _ _ _ _ 7 
\end{taskexample}

Above, the three-token key \texttt{(8,2,5)} has value \texttt{9}, and key \texttt{(3,7,0)} has value \texttt{4}. Key \texttt{(6,1,8)} has value \texttt{2}, and key \texttt{(9,4,3)} has value \texttt{7}. For each key, the model must retrieve its corresponding value from the mapping learned during training. 

In our experiments, the fixed mapping contains 10,000 keys, each consisting of 3 tokens. Each sequence contains 220 tokens.

\subsection{Counting}
\label{sec:counting_task_explanation}

\subsubsection{Counting}

Counting tests whether a model can determine how many times a queried token occurs in the source sequence. The query appears after the separator and is excluded from the count. For example:

\begin{taskexample}[taskletsgreen]
x: 3 4 7 8 4 6 3 5 4 1 | 4 3 3 2 7 1
y: _ _ _ _ _ _ _ _ _ _ _ _ 3 _ 2 _ 1
\end{taskexample}

Above, token \texttt{4} appears three times in the source sequence (before the `\texttt{|}' token). The token \texttt{3} appears twice, and token \texttt{7} appears once. When queried with \texttt{4}, the model must return \texttt{3}. When queried with \texttt{3}, it must return \texttt{2}. When queried with \texttt{7}, it must return \texttt{1}. The other tokens, \texttt{8}, \texttt{6}, \texttt{5}, and \texttt{1}, contribute to none of these counts. The model must track several counts across the same source sequence, including occurrences separated by other tokens.

We use a sequence-length setting of 128, with 124 source tokens and four query tokens per example. Adding the separator and four count answers gives a total length of 133 tokens. No token occurs more than 50 times in the source sequence.

\subsubsection{Sorting}

Sorting requires the model to arrange the distinct tokens according to how often they occur in the source sequence. After the separator, the model must output each distinct token once, starting with the most frequent and ending with the least frequent. For example:

\begin{taskexample}[taskletsgreen]
x: 9 2 7 2 5 9 2 7 9 2 7 2 2 7 | 2 7 9 5
y: _ _ _ _ _ _ _ _ _ _ _ _ _ _ _ 2 7 9 5
\end{taskexample}

Before the `\texttt{|}' token, \texttt{2} occurs six times, \texttt{7} four times, \texttt{9} three times, and \texttt{5} once. The order is thus \texttt{(2,7,9,5)}.

For this task, each source sequence contains 128 tokens drawn from 10 distinct symbols. Including the separator and the ordered list of symbols, the complete sequence contains 139 tokens.

\subsection{Stacks}
\label{sec:stack_task_explanation}

\subsubsection{Stack Operations}

To implement a stack, an architecture must learn to push or insert values onto the top and, when instructed to pop a value, must always remove the most recently added value still on the stack first. Here, a push instruction, denoted by \texttt{i} (for ``insert''), places the number immediately after it onto the top of the stack. A pop instruction, denoted by \texttt{o}, requires the architecture layer to remove any value remaining at the top and return it. For example:

\begin{taskexample}[taskletsteal]
x: i 8 i 3 i 6 o i 9 o o i 7 o o i 5
y: _ _ _ _ _ _ _ 6 _ _ 9 3 _ _ 7 8 _
\end{taskexample}

The first three insertions build the stack \texttt{(8,3,6)}, with \texttt{6} on top. The first pop returns \texttt{6}. The remaining pops return \texttt{(9,3,7,8)}, with \texttt{9} and \texttt{7} inserted before their respective pops.

We use input sequences of 100 tokens to specify push and pop operations on a single stack.

\subsubsection{Multi Stack Operations}

Multi-stack operations extend stack operations to include several independent stacks. Each instruction identifies the stack on which it acts. The example below uses two stacks, identified by \texttt{0} and \texttt{1}.

\begin{taskexample}[taskletsteal]
x: i 0 7 i 0 4 i 1 9 o 0 o 1 o 0 i 1 6
y: _ _ _ _ _ _ _ _ _ _ _ 4 _ 9 _ 7 _ _
\end{taskexample}

After the first three insertions, stack \texttt{0} contains \texttt{(7,4)}, while stack \texttt{1} contains \texttt{9}. The first pop returns \texttt{4} from stack \texttt{0}, even though \texttt{9} was inserted more recently into stack \texttt{1}. The next two pops return \texttt{9} and \texttt{7}, from stacks \texttt{1} and \texttt{0} respectively.

Although this example uses two stacks, the dataset uses four. If each stack has the same depth, multi-stack implementation requires tracking more information than single-stack implementation. However, if the total number of stored values is held fixed, distributing those values across more stacks reduces the average depth per stack.

For this task, we use 4 independent stacks, with each input sequence containing 113 tokens.

\subsection{Formal Languages}
\label{sec:formallanguages_task_explanation}

\subsubsection{Dyck Language}

Dyck-language recognition tests whether a sequence of brackets is balanced and correctly nested. Each closing bracket must match an earlier opening bracket, and no opening brackets may remain unmatched at the end. The model returns \texttt{1} for a valid sequence and \texttt{0} otherwise. For example, a valid sequence is:

\begin{taskexample}[taskletslavender]
x: ( ( ( ) ) ( ) ) ( ) ( ( ) ( ) ) | 1
y: _ _ _ _ _ _ _ _ _ _ _ _ _ _ _ _ _ 1
\end{taskexample}

For comparison, an invalid sequence is:

\begin{taskexample}[taskletslavender]
x: ( ( ( ) ) ( ) ) ) ( ( ( ) ( ) ) | 0
y: _ _ _ _ _ _ _ _ _ _ _ _ _ _ _ _ _ 0
\end{taskexample}

The first sequence is valid. In the second, the ninth bracket closes without a matching opening bracket, and the tenth opens without a matching closing bracket. Swapping these two brackets could be one way to achieve a valid sequence.

In our experiments, the sequence length is 130 tokens (128 tokens before the separator).

\subsubsection{$a^nb^nc^n$ Language}

This task tests whether a sequence contains equally many \texttt{a}, \texttt{b}, and \texttt{c} tokens, arranged in three consecutive blocks in that order. The model returns \texttt{1} when the sequence satisfies these conditions and \texttt{0} otherwise. For example:

\begin{taskexample}[taskletslavender]
x: _ _ a a a a a b b b b b c c c c c | 1
y: _ _ _ _ _ _ _ _ _ _ _ _ _ _ _ _ _ _ 1
\end{taskexample}

\begin{taskexample}[taskletslavender]
x: a a a a a a a b b b b c c c c c c | 0
y: _ _ _ _ _ _ _ _ _ _ _ _ _ _ _ _ _ _ 0
\end{taskexample}  

For this task, sequences have variable lengths and are padded on the left such that the total length of the portion before the separator is 128 tokens.

\subsubsection{Flip Flop Language Modeling}

In flip-flop language modelling, a write instruction, denoted by \texttt{w}, stores the following value. An ignore instruction, denoted by \texttt{i}, leaves the stored value unchanged, while a read instruction, denoted by \texttt{r}, requires the model to return it. In the dataset format below, each read target appears one position after \texttt{r} in \texttt{y}, and read values are not included in \texttt{x}.

\begin{taskexample}[taskletslavender]
x: w 7 i 3 r w 4 i 9 i 2 r w 8 i 5 r w 6
y: _ _ _ _ _ 7 _ _ _ _ _ _ 4 _ _ _ _ 8 _
\end{taskexample}

Above, the three reads return \texttt{(7,4,8)}, each matching the most recent write. The values following \texttt{i} do not change the stored value. The final \texttt{(w,6)} stores \texttt{6}.

In our experiments, each input sequence contains 111 tokens comprising write, ignore, and read instructions and accompanying values.

\subsection{State tracking}  
\label{sec:statetracking_task_explanation}

\subsubsection{Modular Addition}

Modular addition tests whether an architecture layer can maintain a running sum in this case modulo \texttt{10} (\(\mathbb{Z}_{10}\)). Starting from zero, the architectural layer must add each new input to the running total and keep the remainder after division by \texttt{10}. For example:

\begin{taskexample}[taskletscoral]
x: 2 7 5 9 2 3 3 7 2 1 5 0 3 1 7 4 2 6
y: _ 2 9 4 3 5 8 1 8 0 1 6 6 9 0 7 1 3
\end{taskexample}

Above, the first input, \texttt{2}, gives a running total of \texttt{2}. Adding \texttt{7} gives \texttt{9}. Adding \texttt{5} then gives \texttt{14}, whose remainder modulo \texttt{10} is \texttt{4}. Each subsequent input updates the accumulated value according to the same rule. 

In the \tokenlyfont{Tasklets} dataset, we use input sequences of 128 tokens for this task.

\subsubsection{Permutation Composition ($A_5$)}

Permutation composition tests whether a model can track the combined effect of successive rearrangements. Each input token represents one of the 60 even permutations of five positions, which form the alternating group \(A_5\). For example, token \texttt{0} represents \texttt{(0,1,2,3,4)}, token \texttt{15} represents \texttt{(1,2,0,3,4)}, and token \texttt{25} represents \texttt{(2,0,3,4,1)}. Starting from the identity permutation, denoted by token \texttt{0}, the model must apply each permutation in the order presented and return the token representing the cumulative result. For example:

\begin{taskexample}[taskletscoral]
x: 15 15 15 1 13  1  1 29 12 3 24 19 15 6
y: _  15 24 0  1 12 13 14 15 3  0 24 28 4
\end{taskexample}

Above, token \texttt{15} is the permutation \texttt{(1,2,0,3,4)}, which can be implemented by taking five items initially arranged as \texttt{(0,1,2,3,4)} and swapping the contents of positions \texttt{0} and \texttt{1}, then those of positions \texttt{1} and \texttt{2}, while leaving positions \texttt{3} and \texttt{4} unchanged. Applying it once gives \texttt{15}. Applying it again gives \texttt{(2,0,1,3,4)}, token \texttt{24}. Applying it a third time returns every item to its original position, giving the identity, token \texttt{0}. The next input, \texttt{1}, is \texttt{(0,1,3,4,2)}, so the composition becomes \texttt{1}. Applying \texttt{13}, which is \texttt{(1,0,3,2,4)}, then gives \texttt{(1,0,2,4,3)}, token \texttt{12}. Unlike modular addition, the order of the inputs matters: starting from the identity, applying \texttt{13} first and \texttt{1} second instead gives \texttt{(1,0,4,3,2)}, token \texttt{14}.

For this task, each input sequence contains 128 permutation tokens.

We provide details on models tested, and evaluation methodology, in Section~\ref{sec:architectures} and ~\ref{sec:experimental_setup} respectively. Results are analysed in Section~\ref{sec:analysis}.

\section{Methodology}
\label{sec:methodology}

\begin{figure*}[!t]
\begin{center}
\includegraphics[width=\textwidth]{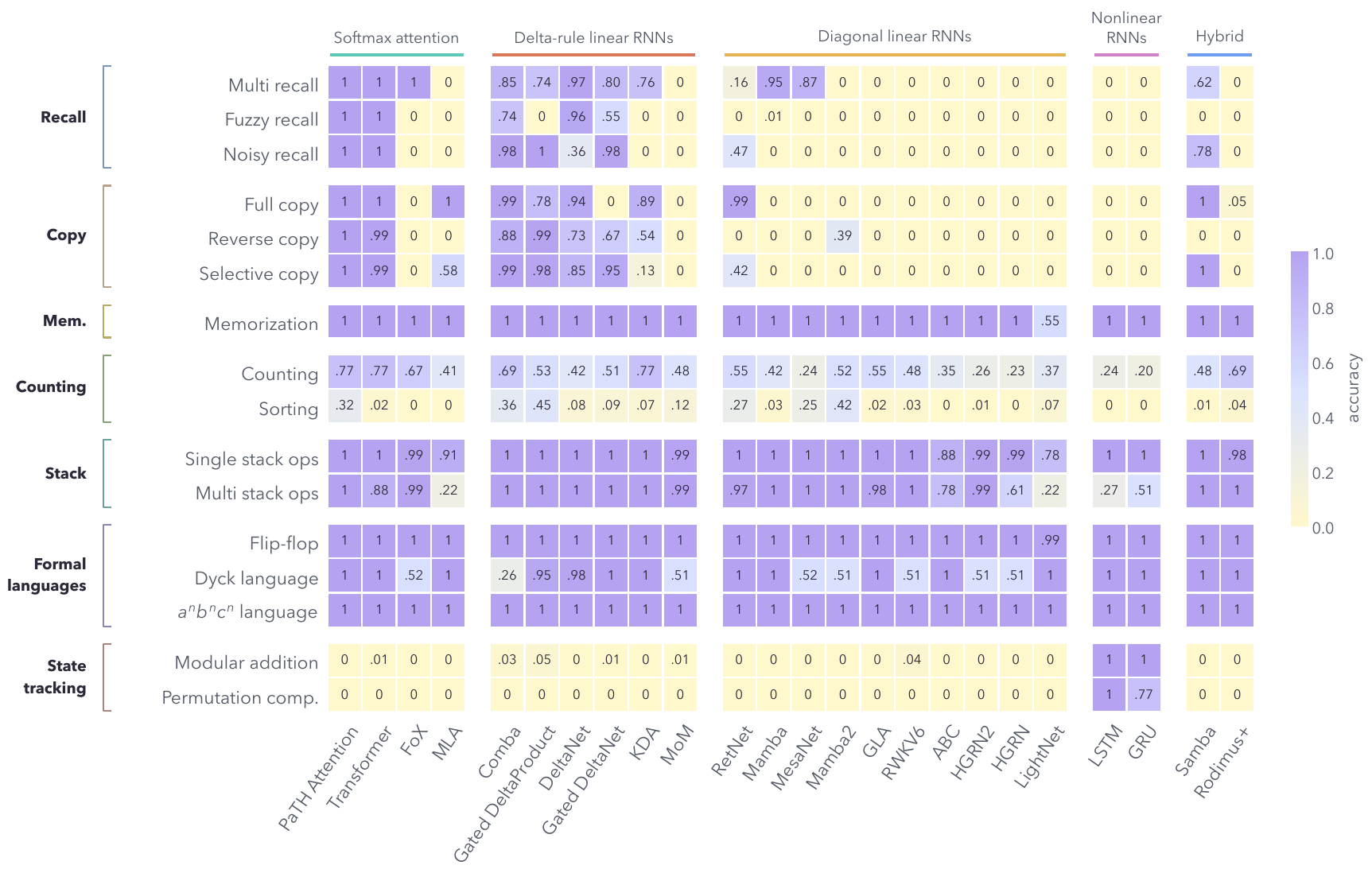}
\end{center}
\caption{Test accuracy of all 24 architectures on the 16 \tokenlyfont{Tasklets} tasks. Columns are grouped by architecture family and sorted within each family by mean accuracy across tasks. Rows are grouped by capability. Each cell reports the test accuracy of the run with the lowest validation loss among the three learning rate runs (Section~\ref{sec:experimental_setup}).}
\label{fig:main_heatmap}
\end{figure*}

We evaluate 24 sequence architectures under a shared training and evaluation setup. Below, we describe the architectures and detailed experimental setup.

\subsection{Evaluated Architectures}
\label{sec:architectures}

We organise the architectures into five families based on their main sequence-mixing operations:

\textbf{Softmax attention:} The original softmax attention \cite{vaswani2017attention,bahdanau2015neural} had attention weights \(a_{ti}=\operatorname{softmax}_{i\leq t}
(q_t^\top k_i/\sqrt{d_k})\) and readout \(o_t=\sum_{i\leq t}a_{ti}v_i\). Our Transformer baseline uses this attention with rotary position embeddings \citep[RoPE;][]{su2024roformer}. We also consider 3 variants:

\begin{itemize}[itemsep=2pt, topsep=3pt, parsep=0pt, partopsep=0pt]

\item FoX \citep{lin2025forgetting} computes a scalar, input-dependent forget gate \(f_t=\sigma(w_f^\top x_t+b_f)\) for each head, and adds the bias \(D_{ti}=\sum_{s=i+1}^{t}\log f_s\) to the attention logits inside the softmax, so that \(a_{ti}=\operatorname{softmax}_{i\leq t}(q_t^\top k_i/\sqrt{d_k}+D_{ti})\).

\item PaTH Attention \citep{yang2025path} computes input-dependent transformations \(H_s=I-\beta_sw_sw_s^\top\), with unit-norm \(w_s\) and \(\beta_s\in(0,2)\), and replaces the query-key dot product \(k_i^\top q_t\) with \(k_i^\top(H_{i+1}\cdots H_t)q_t\). RoPE is the special case in which every \(H_s\) is the same fixed rotation, so PaTH offers an input-dependent alternative to RoPE.

\item MLA \citep{deepseekai2024deepseekv2} jointly compresses keys and values into a low-dimensional latent vector \(c_t^{KV}\), from which each head's keys and values are projected back up, reducing the key-value cache at inference. Position is encoded by a small separate part of each query and key that carries RoPE.

\end{itemize}

\textbf{Delta-rule linear RNNs:} These build on the associative-memory formulation of linear attention, in which a matrix-valued state accumulates key-value associations, \(S_t=S_{t-1}+v_tk_t^\top\), and is read out with \(o_t=S_tq_t\) \citep{katharopoulos2020transformers}. They adapt this state update in various ways:

\begin{itemize}[itemsep=2pt, topsep=3pt, parsep=0pt, partopsep=0pt]

\item DeltaNet \citep{schlag2021linear} addresses a limitation of the additive linear-attention update which keeps writing new key-value associations without removing outdated ones. The delta rule retrieves the value currently stored for \(k_t\), \(S_{t-1}k_t\), and replaces it with an interpolation between this retrieved value and the incoming value \(v_t\), weighted by an input-dependent \(\beta_t\). This gives \(S_t=S_{t-1}+\beta_t(v_t-S_{t-1}k_t)k_t^\top=S_{t-1}(I-\beta_tk_tk_t^\top)+\beta_tv_tk_t^\top\).

\item Gated DeltaNet \citep{yang2025gated} then introduces an input-dependent scalar forget gate \(\alpha_t\in(0,1)\) that scales the prior state, giving \(S_t=\alpha_tS_{t-1}(I-\beta_tk_tk_t^\top)+\beta_tv_tk_t^\top\). Unlike \(\beta_t\), which controls how strongly the value associated with the current key is replaced by the incoming value, \(\alpha_t\) decays the entire previous memory state, allowing accumulated information to be rapidly cleared when outdated or irrelevant.

\item KDA \citep{kimiteam2025kimilinearexpressiveefficient} replaces Gated DeltaNet's scalar forget gate with an input-dependent vector \(\bm{\alpha}_t\), giving \(S_t=S_{t-1}\operatorname{Diag}(\bm{\alpha}_t)(I-\beta_tk_tk_t^\top)+\beta_tv_tk_t^\top\). This assigns a separate decay factor to each key-feature channel of the previous memory state, allowing some channels to retain information while others forget more rapidly.

\item We also evaluate Gated DeltaProduct \citep{siems2025deltaproduct}, which applies \(n_h\) delta-rule updates per token; Comba \citep{hu2025comba}, which replaces Gated DeltaNet's transition \(\alpha_t(I-\beta_tk_tk_t^\top)\) with \(\alpha_t-\tilde{\beta}_tk_tk_t^\top\) and reads out with \(o_t=S_t(q_t-dk_t)\) for a scalar \(d\); and MoM \citep{du2026mom}, which routes each token to its top-\(k\) memories out of several, each updated with the Gated DeltaNet rule, plus a shared memory.

\end{itemize}

\newpage
\textbf{Diagonal linear RNNs:} These models, including diagonal SSMs, update their memory as \(h_t=D_th_{t-1}+u_t\), where the diagonal matrix \(D_t\) scales each state coordinate independently and \(u_t\) adds new input. In SSMs, this structure supports efficient computation and parallelisation \citep{gupta2022diagonal,gu2022parameterization,smith2023simplified}, but can limit state tracking expressivity \citep{shakerinava2026expressive,fujii2026noncommutative}. We evaluate the following variants across all seven capability categories:

\begin{itemize}[itemsep=2pt, topsep=3pt, parsep=0pt, partopsep=0pt]

\item RetNet uses a fixed scalar decay per head \citep{sun2023retentive}, whereas GLA uses an input-dependent vector gate \(\bm{\alpha}_t\) to assign separate decay factors to key-feature channels \citep{yang2024gatedlinear}.

\item Mamba \citep{gu2024mamba} also uses input-dependent diagonal transitions, \(\exp(\Delta_tA)\), where \(A\) is a learned diagonal matrix and the step size \(\Delta_t\) depends on the input. Its input-dependent parameters \(B_t\) and \(C_t\) play roles analogous to keys and queries. Mamba2 \citep{dao2024transformers} restricts the transition to a scalar decay per head, \(a_t=\exp(\Delta_tA)\) with a learned scalar \(A\), giving \(S_t=a_tS_{t-1}+\Delta_tx_tB_t^\top\).

\item HGRN \citep{qin2023hierarchically} is a variant of gated LRNNs \citep{martin2018parallelizing} with a vector state (optionally complex-valued) and a modified forget gate. HGRN2 \citep{qin2024hgrn2} expands this vector state into a matrix-valued state.

\item MesaNet \citep{von2025mesanet} maintains two states, \(G_t=\gamma_tG_{t-1}+\beta_tv_tk_t^\top\) and \(H_t=\gamma_tH_{t-1}+\beta_tk_tk_t^\top\), with scalar gates \(\gamma_t\) and \(\beta_t\). Its readout, \(o_t=G_t(H_t+\Lambda)^{-1}q_t\), applies regularised least-squares regression from keys to values.

\item ABC \citep{peng2022abc} stores keys and values in a fixed number of memory slots and retrieves information through softmax attention over these slots.

\item We also test RWKV-6 \citep{peng2024eagle} and LightNet \citep{qin2026lightnet}.

\end{itemize}

\textbf{Nonlinear RNNs:} These update their recurrent states through nonlinear functions of the previous state and current input. We evaluate LSTMs \citep{lstm6795963,gers2000learning} and GRUs \citep{cho2014learning}, whose gates and updated states depend nonlinearly on the current input and the previous hidden state.

\textbf{Hybrid architectures:} These combine different sequence-mixing mechanisms within one model. We evaluate Samba \citep{ren2025samba}, which interleaves Mamba layers \citep{gu2024mamba} with sliding-window attention \citep{beltagy2020longformer}, combining recurrent compression of the sequence with direct access to recent token representations.
We also evaluate Rodimus+ \citep{he2025rodimus}, which follows each Rodimus layer, a diagonal linear RNN with input-dependent decay, with sliding-window shared-key attention, in which all heads share one key.

\subsection{Experimental Setup}
\label{sec:experimental_setup}

\textbf{Code repository \& architecture implementation.} For training all models on each architecture, we use and extend the flash-linear-attention (FLA) \citep{yang2024fla} and Flame \citep{yang2025flame} repositories. The FLA repository contains implementations of all architectures evaluated in this paper, except for GRU and LSTM, which we implemented ourselves.

\textbf{Training hyperparameters.} Table~\ref{tab:training-hparams} in Appendix~\ref{sec:appendix-a} lists all training hyperparameters. We began with hyperparameters that prior studies had successfully used for particular architecture-task combinations, such as a Transformer on multi-query recall, and confirmed that training converged with these settings. We then applied each architecture's settings to the other tasks, sweeping the learning rate over $\{10^{-4},\,10^{-3},\,10^{-2}\}$. For each architecture-task combination, we report the performance of the best run from this sweep.

Each model has approximately 150,000 - 200,000  parameters. Importantly, the state sizes are smaller than the task sequence lengths. For matrix-valued states, we adjust key and value dimensions to obtain comparable state sizes. Full configurations are provided in Tables~\ref{tab:model-configs} and~\ref{tab:model-specific} in Appendix~\ref{sec:appendix-a}. 

\textbf{Evaluation.} For each task, we hold out $3{,}000$ test samples in the same format as the training data. We obtain predictions through a single forward pass per sample, as opposed to autoregressively.    

We measure exact-match accuracy against the ground truth \texttt{y}, requiring all non-masked positions within a sample to match for that sample to count as correct. Accuracy is averaged over test samples, except for memorization, stack operations, and flip-flop language, where it is averaged over individual non-masked tokens across the test set.

\textbf{Hardware and training time.} One training run of a particular architecture, task, and learning rate combination takes $\sim$7 minutes to train on a single Nvidia GeForce RTX 4090 card. With 24 architectures evaluated on 16 tasks swept over 3 learning rates, totaling 1,152 runs, this equates to $\sim$134 GPU hours on the RTX 4090.

\section{Analysis}
\label{sec:analysis}

\textbf{Recall.} The results show a clear advantage in favor of softmax attention, while some delta-rule linear RNNs perform closely behind.  One reason why delta-rule linear RNNs outperform architectures with diagonal linear RNN could be the ability of the delta rule to directly store and manage key-value associations \citep{schlag2021linear}. In contrast, diagonal linear RNNs parameterise their state transitions using scalar or vector valued scaling factors. Nonlinear RNNs struggle to retrieve associations from the context, consistent with previously reported difficulties on associative recall \citep{graves2014neuralturingmachines}.

\textbf{Copying.} This task closely follows recall as it is a form of sequential recall without queries. Our benchmarking shows an advantage for standard Transformers and several delta-rule architectures over the evaluated SSMs on copying. The evaluated nonlinear RNNs are also disadvantaged here. The advantage over SSMs is consistent with \citet{jelassi2024repeat}.

\textbf{Memorisation.} Memorisation requires learning a fixed mapping during training and retrieving the corresponding values. Almost all evaluated architectures perform well on this task. It therefore provides a useful sanity check for learning fixed associations.

\textbf{Counting.} Both counting \&  sorting tasks remain challenging for virtually all architectures, with sorting being harder than counting. Counting requires maintaining occurrence counts across the sequence, while sorting additionally requires comparing these counts to determine the output order. 

\textbf{Stacks.} Most architectures perform well on single-stack operations, but multi-stack accuracy varies more across architectures. Attention provides direct access to earlier instructions, while recurrent architectures can retain relevant stack information in their evolving states. Previous work has also reported high stack accuracy at short sequence lengths \citep{kimiteam2025kimilinearexpressiveefficient}.

\textbf{Formal languages.} Most architectures in each category learn to recognize formal languages within their expressive capabilities, such as the Dyck language (CFG) and $a^nb^nc^n$ (CS). This is known from previous literature \citep{deletang2023neural}. Our results agree with this.

\textbf{State tracking.} Nonlinear RNNs show a clear advantage on modular addition and permutation composition, demonstrating that strong recall and copying performance does not imply strong state tracking. Prior work also establishes different expressivity bounds for nonlinear RNNs, linear RNN variants, and Transformers \citep{merrill2026parallelizable}.

\section{Conclusion}
\label{sec:conclusion}

We introduced \tokenlyfont{Tasklets}, a computationally inexpensive benchmark spanning seven task categories: recall, copying, memorisation, counting, stack operations, formal languages, and state tracking, totaling 16 synthetic tasks. Our evaluation of 24 architectures shows that most diagonal linear RNNs struggle with recall and copying, while attention and some delta-rule linear RNNs perform well on these tasks. Both linear RNN families and the evaluated softmax attention models nevertheless achieve near zero accuracy on the state tracking tasks, where nonlinear RNNs perform substantially better. These differences highlight the value of testing architectural capabilities on multiple task variants simultaneously, but under the same shared setup, for fair across-the-board comparison. With openly available task implementations and evaluation code, \tokenlyfont{Tasklets} supports reproducible evaluation of new layer designs before committing to larger scale experiments.






\bibliography{references}
\bibliographystyle{icml2026}

\newpage
\appendix

\onecolumn
\section{Experimental Setup Details}
\label{sec:appendix-a}

We present the details of our experimental setup that we used to evaluate the set of architectures in this paper.

%

\begin{table}[h]
\centering
\small
\caption{Training hyperparameters used for the training runs for each architecture-task combination. Each architecture-task combination has 3 training runs swept over 3 learning rates, and the run with the lowest validation loss is evaluated on the test set. Early stopping is done when validation loss $\le 5 \times 10^{-4}$ for 2 consecutive validations.}
\label{tab:training-hparams}
\begin{tabular}{ll}
\toprule
\textbf{Hyperparameter} & \textbf{Value} \\
\midrule
Optimizer                     & AdamW ($\beta_1 = 0.9$, $\beta_2 = 0.95$, $\epsilon = 10^{-15}$) \\
Weight decay                  & 0.1 \\
Learning rate            & $\{10^{-4},\, 10^{-3},\, 10^{-2}\}$ \\
Learning rate schedule        & Linear warmup with cosine decay \\
Warmup steps                  & 1{,}250 \\
Minimum learning rate ratio   & 0.1 \\
Training steps (maximum)      & 12{,}500 \\
Batch size                    & 128 \\
Max. sequence length          & 1{,}024 tokens \\
Gradient clipping             & 1.0 \\
Precision                     & bfloat16 mixed precision \\
Random seed                   & 42 \\
\bottomrule
\end{tabular}
\end{table}

\vspace{2em}
\begin{table}[H]
\centering
\small
\caption{Model size configurations. All models use a vocabulary of 292 tokens and pre-norm
residual blocks. $d_{\text{model}}$ is the hidden size i.e. the residual stream and embedding size while
$d_k$ and $d_v$ are the equivalent per-head key and value state dimensions of the sequence-mixing layer.
``SwiGLU'' denotes a SwiGLU MLP with intermediate size 256. Parameter counts include the input embedding and output head. Architecture-specific settings are detailed in
Table~\ref{tab:model-specific}.}
\label{tab:model-configs}
\begin{tabular}{lccccccr}
\toprule
\textbf{Model} & \textbf{Layers} & $\boldsymbol{d_{\text{model}}}$ & \textbf{Heads} & $\boldsymbol{d_k}$ & $\boldsymbol{d_v}$
  & \textbf{MLP} & \textbf{\# Params} \\
\midrule
\multicolumn{8}{l}{\textit{Softmax attention}} \\
Transformer        & 2 & 64 & 2 & 32 & 32 & SwiGLU & 168{,}768 \\
FoX                & 2 & 64 & 2 & 32 & 32 & SwiGLU & 169{,}028 \\
PaTH               & 2 & 64 & 2 & 32 & 32 & SwiGLU & 177{,}604 \\
MLA                & 2 & 64 & 2 & 16 & 8  & SwiGLU & 143{,}728 \\
\midrule
\multicolumn{8}{l}{\textit{Delta-rule linear RNNs}} \\
DeltaNet           & 2 & 64 & 2 & 32 & 32  & SwiGLU & 170{,}624 \\
Gated DeltaNet     & 2 & 64 & 2 & 32 & 32  & SwiGLU & 179{,}080 \\
KDA                & 2 & 64 & 2 & 32 & 32  & SwiGLU & 187{,}268 \\
Gated DeltaProduct & 2 & 64 & 2 & 32 & 32  & SwiGLU & 196{,}744 \\
Comba              & 2 & 64 & 2 & 32 & 64  & SwiGLU & 204{,}240 \\
MoM                & 2 & 64 & 2 & 16 & 16  & SwiGLU & 193{,}128 \\
\midrule
\multicolumn{8}{l}{\textit{Diagonal linear RNNs}} \\
RetNet             & 2 & 64 & 2 & 32 & 64  & SwiGLU & 201{,}664 \\
GLA                & 2 & 64 & 2 & 16 & 32  & SwiGLU & 171{,}968 \\
HGRN               & 2 & 64 & -- & \multicolumn{2}{c}{64 (element-wise)} & SwiGLU & 169{,}024 \\
HGRN2              & 2 & 64 & 2 & 32 & 32  & SwiGLU & 169{,}024 \\
RWKV6              & 2 & 64 & 2 & 16 & 32  & channel mix & 159{,}360 \\
ABC                & 2 & 64 & 2 & 16 & 32  & SwiGLU & 172{,}928 \\
LightNet           & 2 & 64 & 2 & 8  & 32  & SwiGLU & 158{,}656 \\
Mamba              & 4 & 64 & -- & \multicolumn{2}{c}{$128 \times 16$ (channels $\times$ state)} & -- & 168{,}256 \\
Mamba2             & 4 & 64 & 4 & 16 & 32  & --     & 148{,}976 \\
MesaNet            & 2 & 64 & 2 & 32 & 32  & SwiGLU & 170{,}504 \\
\midrule
\multicolumn{8}{l}{\textit{Nonlinear RNNs}} \\
LSTM               & 4 & 64 & -- & \multicolumn{2}{c}{64 (hidden)} & -- & 187{,}200 \\
GRU                & 4 & 64 & -- & \multicolumn{2}{c}{64 (hidden)} & -- & 153{,}920 \\
\midrule
\multicolumn{8}{l}{\textit{Hybrid architectures}} \\
Rodimus+           & 2 & 64 & 1 & 8  & 128 & SwiGLU & 213{,}728 \\
Samba              & 2 & 64 & 2 & \multicolumn{2}{c}{see Table~\ref{tab:model-specific}} & SwiGLU & 180{,}416 \\
\bottomrule
\end{tabular}
\end{table}

\begin{table}[!t]
\centering
\small
\renewcommand{\arraystretch}{1.2}
\caption{Architecture-specific settings of the sequence mixers. ``Short conv.'' denotes a depthwise causal convolution on the mixer inputs with kernel size 4 unless stated otherwise. Low-rank gates and
projections use the stated bottleneck rank.}
\label{tab:model-specific}
\begin{tabularx}{\linewidth}{@{}lX@{}}
\toprule
\textbf{Model} & \textbf{Settings} \\
\midrule
\multicolumn{2}{@{}l}{\textit{Softmax attention}} \\

Transformer &
Multi-head attention with RoPE ($\theta = 10^4$), without QK-normalisation or attention bias. \\

FoX &
A data-dependent scalar forget gate per head, without an output gate. \\

PaTH &
Householder position encoding with a rank-32 projection bottleneck and short conv.\ (kernel 3) on $\mathbf{w}$, without a forget gate. \\

MLA &
Query LoRA rank 8 and KV LoRA rank 16, with query/key head dimension 16 (8 non-positional and 8 with RoPE, $\theta = 10^4$) and value head dimension 8. \\

\midrule
\multicolumn{2}{@{}l}{\textit{Delta-rule linear RNNs}} \\

DeltaNet &
Data-dependent $\beta_t$ with SiLU and $L_2$ normalisation of queries and keys, short conv., and output normalisation, without an output gate. \\

Gated DeltaNet &
A scalar decay gate $\alpha_t$ per head, with short conv.\ and an output gate. \\

KDA &
A channel-wise decay gate and an output gate, each with a rank-32 bottleneck, together with short conv. \\

Gated DeltaProduct &
Two Householder steps per token ($n_h = 2$) and a forget gate, with short conv.\ and an output gate, but no negative eigenvalues. \\

Comba &
Inner decay and output correction ($d = 1.0$), with short conv.\ and an output gate. \\

MoM &
Four memories with top-2 routing plus a shared memory, using Gated DeltaNet updates with short conv.\ and an output gate. The load-balancing loss has weight 0.01. \\

\midrule
\multicolumn{2}{@{}l}{\textit{Diagonal linear RNNs}} \\

RetNet &
Fixed exponential decay per head, with an output gate and no short conv. \\

GLA &
A channel-wise key gate with a rank-16 bottleneck, together with an output gate and no short conv. \\

HGRN &
Element-wise recurrence with a layer-wise lower bound on the forget gate. \\

HGRN2 &
An outer-product state with expansion ratio 32 and a layer-wise lower bound on the forget gate. \\

RWKV6 &
Data-dependent decay and token shift with rank-16 and rank-8 bottlenecks, respectively, plus a channel-mixing FFN of width 256 with squared-ReLU activation. \\

ABC &
16 memory slots per head with data-dependent slot weights and RoPE on queries and keys. \\

LightNet &
An output gate with a rank-8 bottleneck. \\

Mamba &
An inner width of 128 (expansion factor 2) and SSM state dimension 16, with convolution kernel size 4 and $\Delta$ rank 4. \\

Mamba2 &
An inner width of 128 (expansion factor 2), with 4 heads of dimension 32, SSM state dimension 16, one group, convolution kernel size 4, and chunk size 16. \\

MesaNet &
An online least-squares readout computed using 30 conjugate-gradient steps, with a lower bound of 0.25 on $\lambda$, short conv., and no output gate. \\

\midrule
\multicolumn{2}{@{}l}{\textit{Nonlinear RNNs}} \\

LSTM &
A single-layer LSTM cell with biases followed by a linear output projection in each block, with no MLP. \\

GRU &
A single-layer GRU cell with biases followed by a linear output projection in each block, with no MLP. \\

\midrule
\multicolumn{2}{@{}l}{\textit{Hybrid architectures}} \\

Rodimus+ &
A Rodimus mixer (inner width 128, state dimension 8, short conv.) followed in each block by shared-key softmax attention (2 heads of dimension 32, RoPE $\theta = 10^4$, window 32 tokens). \\

Samba &
A Mamba first layer (inner width 128, SSM state dimension 4, convolution kernel size 4) followed by a sliding-window attention second layer (2 heads of dimension 32, window 128, RoPE $\theta = 10^4$). Both layers use a SwiGLU MLP. \\

\bottomrule
\end{tabularx}
\end{table}

\end{document}